\documentclass[final]{cvpr}

\usepackage{times}
\usepackage{epsfig}
\usepackage{graphicx}
\usepackage{amsmath}
\usepackage{amssymb}
\usepackage{makecell}

\usepackage{bm} 
\usepackage{multirow} 
\usepackage{hhline} 
\usepackage{array} 
\newcolumntype{M}[1]{>{\centering\arraybackslash}m{#1}} 
\usepackage[utf8]{inputenc}
\usepackage[english]{babel}
\usepackage[dvipsnames]{xcolor}
\definecolor{myred}{RGB}{242,70,153}
\definecolor{mygreen}{RGB}{153,206,152}
\definecolor{myblue}{RGB}{83,70,232}

\usepackage{anyfontsize} 
\usepackage{mathtools} 

\newlength{\textfloatsepsave} 
\usepackage{algorithmicx}
\usepackage[ruled]{algorithm}
\usepackage[noend]{algpseudocode}

\usepackage{varwidth}
\usepackage{lipsum,etoolbox}
\usepackage{graphics}
\usepackage{amssymb}
\usepackage{pifont}
\newcommand{\cmark}{\ding{51}}%
\newcommand{\xmark}{\ding{55}}%

\usepackage[pagebackref=true,breaklinks=true,colorlinks,bookmarks=false]{hyperref}

\def\confYear{CVPR 2021}

\begin{document}

\title{Meta Batch-Instance Normalization for Generalizable Person Re-Identification}

\author{Seokeon Choi  \quad Taekyung Kim \quad Minki Jeong \quad Hyoungseob Park \quad Changick Kim\\
Korea Advanced Institute of Science and Technology, Daejeon, Republic of Korea\\
{\tt\small $\{$seokeon, tkkim93, rhm033, hyoungseob, changick$\}$@kaist.ac.kr}
}

\maketitle

\begin{abstract}
\vspace{-0.10cm}
Although supervised person re-identification (Re-ID) methods have shown impressive performance, they suffer from a poor generalization capability on unseen domains. Therefore, generalizable Re-ID has recently attracted growing attention. Many existing methods have employed an instance normalization technique to reduce style variations, but the loss of discriminative information could not be avoided. In this paper, we propose a novel generalizable Re-ID framework, named Meta Batch-Instance Normalization (MetaBIN). Our main idea is to generalize normalization layers by simulating unsuccessful generalization scenarios beforehand in the meta-learning pipeline. To this end, we combine learnable batch-instance normalization layers with meta-learning and investigate the challenging cases caused by both batch and instance normalization layers. Moreover, we diversify the virtual simulations via our meta-train loss accompanied by a cyclic inner-updating manner to boost generalization capability. After all, the MetaBIN framework prevents our model from overfitting to the given source styles and improves the generalization capability to unseen domains without additional data augmentation or complicated network design. Extensive experimental results show that our model outperforms the state-of-the-art methods on the large-scale domain generalization Re-ID benchmark and the cross-domain Re-ID problem. The source code is available at: \href{https://github.com/bismex/MetaBIN}{https://github.com/bismex/MetaBIN}.

\end{abstract}
\vspace{-0.25cm}
\section{Introduction}
\label{sec:Introduction}
\vspace{-0.10cm}

Person re-identification (Re-ID) aims to identify a specific person across non-overlapping cameras under various viewpoints and locations. Re-ID has attracted extensive research attention thanks to its practical importance in surveillance systems. With the development of deep Convolution Neural Networks (CNNs), person Re-ID methods~\cite{zhou2020online, zhuang2020rethinking, zhang2020relation, chen2020salience, liu2020unity} have achieved remarkable performance in a supervised manner, where a model is trained and tested on separated splits of the same dataset. However, this supervised approach is hardly applicable in practice due to expensive labeling costs and also suffers from severe performance degradation on an unseen target domain. For resolving this problem, unsupervised domain adaptation (UDA) methods~\cite{zhai2020ad, wang2020smoothing, chen2019instance, fu2019self, zhang2019self, liu2019adaptive} have been introduced, which adapt a Re-ID model from a labeled source domain to an unlabeled target domain. The UDA approach is more practical than the supervised approach, but data collection is still required for updating the model on the target domain.

\begin{figure}[t]
\begin{center}
\vspace{-0.0cm}
\includegraphics[width=1.0\linewidth]{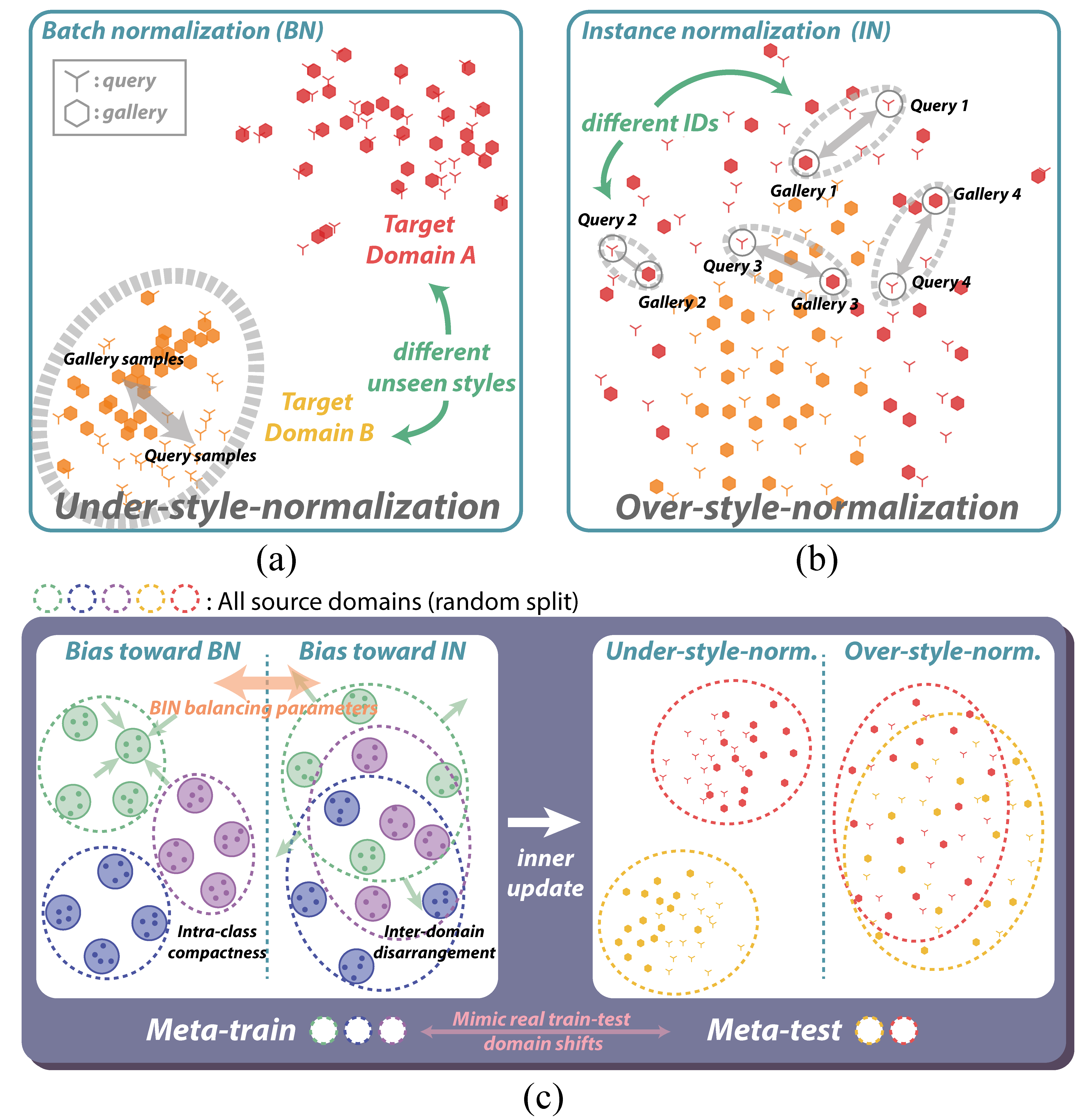}
\end{center}
\vspace{-0.20cm}
\caption{
Illustration of unsuccessful generalization scenarios and our framework. (a) \textit{Under-style-normalization} happens when the trained BN model fails to distinguish identities on unseen domains. (b) \textit{Over-style-normalization} happens when the trained IN model removes even ID-discriminative information. (c) Our key idea is to generalize BIN layers by simulating the preceding cases in a meta-learning pipeline. By overcoming the harsh situations, our model learns to avoid overfitting to source styles.
}
\vspace{-0.20cm}
\label{fig:concept}
\end{figure}

Beyond the concept of UDA, the task of domain generalization (DG) is more plausible for real-world applications since it does not require any target images to train a model. Since Finn \etal~\cite{finn2017model} proposed the Model-Agnostic Meta-Learning (MAML) scheme for few-shot learning and reinforcement learning, several MAML-based methods~\cite{balaji2018metareg, li2017learning} have been investigated to solve the DG problem. This approach enables a DG model to achieve a good generalization capability by dividing multiple source domains into meta-train and meta-test domains to mimic real train-test domain shifts. However, most DG methods~\cite{balaji2018metareg, li2017learning} assume a homogeneous environment, where the source and target domains share the same label space, and they are designed for a classification task. In contrast, the task of domain generalization for person re-identification (DG Re-ID) deals with different label spaces between source and target domains for a retrieval task. Thus, it is difficult to obtain good performance when the existing DG methods are directly applied to DG Re-ID.

To this end, the recent DG Re-ID methods~\cite{jia2019frustratingly, jin2020style, zhou2019learning} have focused on a combination of batch normalization (BN)~\cite{ioffe2015batch} and instance normalization (IN)~\cite{ulyanov2016instance}. Jia \etal~\cite{jia2019frustratingly} adopted this idea to Re-ID by inserting IN after the residual connection in the specific layers, inspired by \cite{pan2018two}. However, this na\"ive approach not only leads to the loss of discriminative information but also requires a careful selection of the locations for applying IN. For another instance, Jin \etal~\cite{jin2020style} designed the Style Normalization and Restitution (SNR) module based on instance normalization and feature distillation. Since this method aims at removing style discrepancy only from the given source domains, it lacks the ability to attenuate the new style of unseen domains sufficiently.

So, how do we design normalization layers to be well generalized for DG Re-ID? To find out the answer, we conducted simple experiments that explore the properties and limitations of BN and IN layers. After training each of the BN and IN models on multiple source domains, we observed the retrieval results on unseen target domains. The BN model strives to learn discriminative information based on the style variations within each mini-batch. However, when samples of unexpected styles are given from unseen target domains, the trained model does not have sufficient ability to distinguish their IDs. We call it \textit{under-style-normalization} in Fig.~\ref{fig:concept}~(a). On the contrary, IN eliminates instance-specific style information using its own statistics. Even though IN can be helpful to remove unseen styles on target domains, it filters out even some discriminative information, as shown in Fig.~\ref{fig:concept}~(b). We call it \textit{over-style-normalization}. Since both normalization methods have limitations in the DG setting, the combination of BN and IN has to be handled carefully.

To deal with the above issues, we propose a novel generalizable Re-ID framework, named Meta Batch-Instance Normalization (MetaBIN), for learning to generalize normalization layers. The key idea is to simulate the unsuccessful generalization scenarios mentioned earlier within a meta-learning pipeline and learn more generalized representation from the virtual simulations, as shown in Fig.~\ref{fig:concept}~(c). For this purpose, we design a batch-instance normalization with learnable balancing parameters between BN and IN. Depending on the balancing parameter's bias toward BN or IN, the DG model suffers from both \textit{under-style-normalization} and \textit{over-style-normalization} scenarios in meta-learning. By overcoming these challenging cases, the normalization layer becomes generalized. Moreover, we intentionally diversify the virtual simulations via our meta-train loss and a cyclic inner-updating manner to effectively boost the generalization capability. Our MetaBIN framework enables to train the DG model equipped with a sufficient generalization capability to novel domains.

Our main contributions can be summarized as follows:
\vspace{-0.15cm}
\begin{itemize}
\item[$\bullet$] We propose a novel generalizable Re-ID framework called MetaBIN. This approach prevents our model from overfitting to the given source styles by generalizing normalization layers via the simulation of unsuccessful generalization scenarios in meta-learning.
\vspace{-0.15cm}

\item[$\bullet$] We diversify the virtual simulations through our meta-train loss and a cyclic inner-updating manner. Both ideas effectively boost generalization capability.
\vspace{-0.15cm}

\item[$\bullet$] We make comprehensive comparisons and achieve state-of-the-art performance on the large-scale domain generalization Re-ID benchmark and the cross-domain Re-ID problem.
\end{itemize}
\vspace{-0.10cm}

\section{Related Work}
\label{Related Work}
\vspace{-0.10cm}

\textbf{Generalizable person re-identification:}
Domain generalizable person re-identification (DG Re-ID) aims to learn a robust model for obtaining good performance on an unseen target domain without additional updates. In recent years, DG Re-ID has attracted a rapidly growing interest owing to its practical applications. Existing methods are categorized into two groups depending on the configuration of source datasets. One is to learn a robust model from a single dataset~\cite{zhou2019learning, jin2020style, liao2019interpretable, sovrasov2020building} and the other is to utilize multiple large-scale datasets~\cite{song2019generalizable, tamura2019augmented, jia2019frustratingly}. In the context of DG Re-ID based on multiple source datasets, Song \etal~\cite{song2019generalizable} constructed the large-scale domain generalization Re-ID benchmark to validate the generalization capability of DG models. They also designed Domain-Invariant Mapping Network (DIMN) to learn a mapping between a person image and its ID classifier. However, the additional mapping network slows the inference speed. Tamura \etal~\cite{tamura2019augmented} suggested a simple selection strategy for data augmentation. While this method is relatively lightweight and easily applicable to other models, it lacks consistency in maintaining generalization capabilities across various unseen styles. For another instance, Jia \etal~\cite{jia2019frustratingly} inserted instance normalization (IN)~\cite{ulyanov2016instance} into each bottleneck in shallow layers to eliminate useless style information, but IN brings about the loss of discriminative information unavoidably. Unlike these methods, we propose an effective framework to generalize a DG model via the simulation of poorly generalized scenarios without an additional test module and data augmentation.

\textbf{Batch-instance normalization:}
Normalization techniques are widely used in most of the deep neural network architectures for better optimization and regularization. In particular, a combination of batch normalization (BN)~\cite{ioffe2015batch} and instance normalization (IN)~\cite{ulyanov2016instance} has recently gained attention as a technique for improving generalization capability to novel domains, which can be divided into non-parametric and parametric methods. The non-parametric methods~\cite{jia2019frustratingly, pan2018two, zhou2019learning} focus on strategies to replace BN with IN for some layers or add IN to specific locations. However, since their performances depend on where BN and IN are mixed, it is difficult to keep task-independent consistency and requires a lot of trial and error. In contrast, the parametric methods concentrate on how to combine BN and IN in a learnable manner. For example, Nam \etal~\cite{nam2018batch} introduced an effective batch-instance normalization layer through a simple training strategy in which BN and IN are balanced with learnable parameters. Although it has improved recognition performance compared to BN, its learnable manner sometimes leads to overfitting the style of given source domains in the DG setting. In addition, some parametric methods have been studied to handle the following tasks: few-shot learning~\cite{bronskill2020tasknorm} and homogeneous DG~\cite{seo2019learning}. However, these methods are rarely applicable for DG Re-ID, because evaluation procedures are different from that of DG Re-ID. In this paper, we propose a novel framework that generalizes normalization layers more effectively to solve the challenging DG Re-ID problem.

\textbf{Model-agnostic meta-learning for DG:}
Model-Agnostic Meta-Learning (MAML)~\cite{finn2017model} is a meta-learning framework for few-shot learning and reinforcement learning. MAML helps find a good initialization of parameters that are sensitive to changes in novel tasks. Recently, the MAML-based approach has been widely applied to various tasks, such as domain generalization~\cite{li2017learning, balaji2018metareg}, object tracking~\cite{wang2020tracking}, and frame interpolation~\cite{choi2020scene}. From the perspective of DG, MLDG~\cite{li2017learning} is a representative method based on MAML, which enables to train models with a good generalization capability to novel domains via the simulation of real train-test domain-shifts. Moreover, Balaji \etal~\cite{balaji2018metareg} trained regularizers in a meta-learning pipeline to achieve good cross-domain generalization. However, since these methods are designed for homogeneous DG, the performance inevitably decreases when these methods are applied directly to DG Re-ID. Inspired by those methods, we propose a novel generalizable Re-ID framework that combines normalization layers with meta-learning.

\section{Proposed Method}
\label{Proposed Method}

\subsection{Problem Formulation}

We begin with a formal description of the domain generalizable person re-identification (DG Re-ID) problem. We assume that we are given $K$ source domains $\mathcal{D}=\{\mathcal{D}_k\}_{k=1}^K$. Each source domain contains its own image-label pairs $\mathcal{D}_k=\{(\bm{x}_i^k,y_i^k)\}_{i=1}^{N_k}$, where ${N_k}$ is the number of images in the source domain ${\mathcal{D}_k}$. Each sample $\bm{x}_i^k \in \mathcal{X}_k$ is associated with an identity label $y_i^k \in \mathcal{Y}_k = \{1,2,\ldots,M_k\}$, where ${M_k}$ is the number of identities in the source domain ${\mathcal{D}_k}$. While all domains share the label spaces in the general homogeneous DG setting as $\mathcal{Y}_i\!=\!\mathcal{Y}_j\!=\!\mathcal{Y}_*$, $\forall{i,\!j}, 1\!\leq\!i,\!j\!\leq\!K$, source and target domains have completely disjoint label spaces in the DG Re-ID setting as $\mathcal{Y}_i\!\neq\!\mathcal{Y}_j\!\neq\!\mathcal{Y}_*$. In other words, this task is an application of heterogeneous DG, such that the number of identities in all source domains can be expressed as $M= \sum_{k=1}^{K} {M_k}$. In the training phase, we train a DG model using the aggregated image-label pairs of all source domains. In the testing phase, we perform a retrieval task on unseen target domains without additional model updating.

\subsection{Batch-Instance Normalization for DG-ReID}
\label{sec:bin}
In line with our goal of DG, we employ an instance normalization technique~\cite{ulyanov2016instance} to generalize well on unseen domains. Similar to \cite{nam2018batch}, we design a batch-instance normalization module as a mixture of BN and IN with learnable balancing parameters for each channel of all normalization layers. Let $\mathbf{x} \in \mathbb{R}^{N \times C \times H \times W}$ be an input mini-batch of a certain layer, where $N$, $C$, $H$, and $W$ denote the batch, channel, height, and width dimensions, respectively. We combine batch normalization (BN) with instance normalization (IN) as follows:
\vspace{-0.15cm}
\begin{equation} 
\fontsize{9.5}{10}\selectfont
\mathbf{y} = \mathbf{\rho} \left( \mathbf{\gamma}_{B} \cdot \hat{\mathbf{x}}_{B} + \mathbf{\beta}_{B} \right) + (1 - \mathbf{\rho}) \left( \mathbf{\gamma}_{I} \cdot \hat{\mathbf{x}}_{I} + \mathbf{\beta}_{I} \right),
\vspace{-0.15cm}
\label{eq:bin}
\end{equation}
\noindent
where the subscripts $B$ and $I$ denote variables with respect to BN and IN, $\hat{\mathbf{x}}$ is the normalized response by mean and variance, $\mathbf{\gamma}, \mathbf{\beta} \in \mathbb{R}^C$ are affine transformation parameters, and $\mathbf{\rho} \in [0,1]^C$ is an additional learnable parameter to balance BN and IN. Note that $\mathbf{\rho}$ is a channel-wise parameter for each normalization layer in a feature extractor. Different from \cite{nam2018batch}, we apply each affine transformation to normalized responses for BN and IN, which allows batch-instance normalization layers to learn various representations.

However, the training mechanism of balancing BN and IN through learnable parameters is a critical issue that can easily overfit the source domain's styles. In other words, the learnable manner forces the balancing parameters to be optimized depending only on styles within source domains. As a result, the normalization layers are unsuccessfully generalized at given unseen target domains, as shown in Fig.~\ref{fig:concept} (a) and (b). To deal with this problem, we develop our normalization model by using a meta-learning pipeline.

\subsection{Meta Batch-Instance Normalization}

\textbf{MetaBIN framework:}
MLDG~\cite{li2017learning} is a representative DG method based on Model-Agnostic Meta-Learning (MAML)~\cite{finn2017model}, which enables to train models with a good generalization capability to novel domains through the virtual simulation of real train-test domain-shifts. Inspired by this, we apply the MAML scheme to the updating process of the balancing parameters to prevent our model from overfitting to the source style. To this end, we separate an episode that updates the balancing parameters from another episode that updates the rest of the parameters, and then perform both episodes alternately at each training iteration. The separation of the base model updating process not only ensures baseline performance but also enables effective learning compared to updating all parameters in MLDG~\cite{li2017learning}.

Formally, we denote a classifier $g_{\phi}(\cdot)$ parameterized by $\phi$ and a feature extractor $f_{\theta}(\cdot)$ parameterized by $\theta = (\theta_f, \theta_\rho)$, where $\theta_\rho$ and $\theta_f$ are the balancing parameters and the remaining parameters of the feature extractor, respectively. In the base model updating process, we update the classifier parameters $\phi$ and the feature extractor parameters $\theta_f$ except the balancing parameters $\theta_\rho$. In the meta-learning process, only the balancing parameters $\theta_\rho$ are updated. The overall methodological flow is summarized in Algorithm \ref{alg:metabin}.

\textbf{Base model updates:} 
In this stage, we update all parameters without the balancing parameters $\theta_\rho$. To this end, we sample a mini-batch $\mathcal{X}_{B}$ by aggregating the labeled images from all source domains $\mathcal{D}$. We adopt two loss functions for updating a base model. First, we use the cross-entropy loss $\mathcal{L}_\text{ce}$ for ID-discriminative learning as follows:
\vspace{-0.25cm}
\begin{equation}
\fontsize{9.5}{10}\selectfont
\mathcal{L}_\text{ce}(\mathcal{X}_{B}; \theta, \phi)=
\frac{1}{N_{B}}\sum_{i=1}^{N_{B}}
l_\text{ce}\Big({g_{\phi}\big({ f_{\theta}({\bm{x}_i})}\big) }, y_i \Big), 
\vspace{-0.20cm}
\label{eq:ce}
\end{equation}
\noindent
where $N_{B}$ denotes the number of samples in a mini-batch $\mathcal{X}_{B}$. In addition, we apply the label-smoothing method~\cite{szegedy2016rethinking} since there are too many identities in aggregated source domains. It helps prevent our model from overfitting to training IDs.  

Most of the Re-ID methods~\cite{choi2020hi, wang2019learning, jin2020style} combine the cross-entropy loss with the triplet loss together for similarity learning. The second loss is expressed as follows:
\vspace{-0.25cm}
\begin{equation} 
\fontsize{9.5}{10}\selectfont
\mathcal{L}_\text{tr}(\mathcal{X}_{B}; \theta) = \!\frac{1}{N_{B}}\! \sum_{i=1}^{N_{B}} \left[  d(\bm{f}_i^a, \bm{f}_i^p) -  d(\bm{f}_i^a, \bm{f}_i^n) + m \right]_+\!,
\vspace{-0.20cm}
\label{eq:trip}
\end{equation}
\noindent
where $\bm{f}_i^a = f_{\theta}(x_i^a)$ indicates the feature vector of an anchor sample $x_i^a$, $d(\cdot, \cdot)$ is the Euclidean distance, $m$ is a margin parameter, and $\left[ z \right]_+ = \max(z, 0)$. For each sample $\bm{f}_i^a$, we select the hardest positive sample $\bm{f}_i^p$ and the hardest negative sample $\bm{f}_i^n$ within a mini-batch in the same way as \cite{hermans2017defense}. The triplet loss helps enhance the intra-class compactness and inter-class separability in the Euclidean space. To maximize the synergy between $\mathcal{L}_\text{ce}$ and $\mathcal{L}_\text{tr}$, we use the BNNeck structure as proposed in \cite{luo2019bag}. The overall loss is formulated as follows:
\vspace{-0.20cm}
\begin{equation}
\fontsize{9.5}{10}\selectfont
\mathcal{L}_\text{base} (\mathcal{X}_{B} ; \theta, \phi) = \mathcal{L}_\text{ce}(\mathcal{X}_{B}; \theta, \phi) + \mathcal{L}_\text{tr}(\mathcal{X}_{B}; \theta).
\vspace{-0.25cm}
\label{eq:base}
\end{equation}
Then we update our base model as follows:
\vspace{-0.20cm}
\begin{equation}
\begin{split}
\fontsize{10.5}{0}\selectfont
(\theta_f, \phi) \leftarrow \big(\theta_f &- \alpha {\large{\nabla_{\theta_f}}} \mathcal{L}_\text{base} (\mathcal{X}_{B} ; \theta_f, \theta_\rho, \phi), \\[-0.05cm]
\phi &- \alpha \nabla_{\phi} \, \mathcal{L}_\text{base}  (\mathcal{X}_{B} ; \theta_f, \theta_\rho, \phi) \big).
\end{split}
\label{eq:base_update}
\end{equation}
We note that the balancing parameters $\theta_\rho$ are not updated.

\textbf{Domain-level sampling and meta-learning:}
To achieve domain generalization, we split source domains $\mathcal{D}$ into meta-train domains $\mathcal{D}_\text{mtr}$ and meta-test domains $\mathcal{D}_\text{mte}$ randomly at each iteration. This separation is to mimic real train-test domain-shifts to generalize our normalization layers. In this way, we inner-update the balancing parameters via the meta-train loss, and then validate the updated model at unseen-like meta-test domains. Next, we meta-update the balancing parameters via the meta-test loss. After all, the balancing parameters and base model parameters are alternately generalized in the whole process.

\begin{algorithm}[t]
   \caption{MetaBIN}
   \label{alg:metabin}
\textbf{Input}: Source domains $\mathcal{D}=\left\{ {{\mathcal{D}_1},{\mathcal{D}_2}, \ldots , {\mathcal{D}_K}} \right\}$, \\
$\text{\quad\quad\,\,\,\,\,}$ pre-trained parameters $\theta_f$, hyperparameters $\alpha$, $\beta$, $\gamma$. \\
\textbf{Output}: Feature extractor $f_\theta (\cdot)$, classifier $g_\phi (\cdot)$
\begin{algorithmic}[1]
\State Initialize parameters $\theta_\rho$, $\phi$
\For{ite \textbf{in} iterations}
\State \textbf{Base model update}: \hfill// Eq.~(\ref{eq:ce})-Eq.~(\ref{eq:base_update})
\State Sample a mini-batch $\mathcal{X}_{B}$ from $\mathcal{D}$.
\State $\mathcal{L}_\text{base} (\mathcal{X}_{B} ; \theta, \phi) = \mathcal{L}_\text{ce}(\mathcal{X}_{B}; \theta, \phi) + \mathcal{L}_\text{tr}(\mathcal{X}_{B}; \theta)$
\State 
\begin{minipage}[t]{0.4\linewidth}
\vspace{-0.75cm}
\begin{align*}
(\theta_f, \phi) \leftarrow \big(\theta_f &- \alpha\, {\large\!{\nabla_{\theta_f}}} \mathcal{L}_\text{base} (\mathcal{X}_{B} ; \theta_f, \theta_\rho, \phi), \\[-0.1cm]
\phi &- \alpha\, {\large\!{\nabla_{\phi}}} \, \mathcal{L}_\text{base}  (\mathcal{X}_{B} ; \theta_f, \theta_\rho, \phi) \big) 
\end{align*}
\end{minipage}
\State \textbf{Domain-level sampling}:
\State Split $\mathcal{D}$ as ($\mathcal{D}_\text{mtr} \cap \mathcal{D}_\text{mte} = \emptyset$, $\mathcal{D}_\text{mtr} \cup \mathcal{D}_\text{mte} = \mathcal{D}$)
\State \textbf{Meta-train}: \hfill// Eq.~(\ref{eq:scatter})-Eq.~(\ref{eq:mtrain_update})
\State Sample a mini-batch $\mathcal{X}_{S}$ from $\mathcal{D}_\text{mtr}$.
\State $\mathcal{L}_\text{mtr}(\mathcal{X}_{S} ; \theta)\!\!=\!\!\mathcal{L}_\text{scat}(\mathcal{X}_{S}; \theta)\!+\! \mathcal{L}_\text{shuf}(\mathcal{X}_{S}; \theta)\!+\!\mathcal{L}_\text{tr}(\mathcal{X}_{S}; \theta)$
\State $\theta_\rho' = \theta_\rho - \beta$  {\large\!$\nabla_{\!\theta_\rho}\!$} $\mathcal{L}_\text{mtr} (\mathcal{X}_{S} ; \theta_f, \theta_\rho)$
\State \textbf{Meta-test}: \hfill// Eq.~(\ref{eq:mtest_update})
\State Sample a mini-batch $\mathcal{X}_{T}$ from $\mathcal{D}_\text{mte}$.
\State $\theta_\rho \leftarrow \theta_\rho - \gamma$  {\large\!$\nabla_{\!\theta_{\rho}}\!$} $\mathcal{L}_\text{tr} (\mathcal{X}_{T} ; \theta_f, \theta'_\rho)$
\EndFor
\end{algorithmic}
\end{algorithm}

\textbf{Meta-train:}
Compared to the previous DG methods~\cite{li2017learning, balaji2018metareg}, we introduce a novel concept to significantly improve generalization capability. We start with an explanation of virtual simulations, as shown in Fig.~\ref{fig:concept} (c). We attempt to simulate unsuccessful generalization scenarios via our meta-train loss by inner-updating the balancing parameters to any space that cannot be explored with the given source domains and general classification loss functions. More specifically, we induce the balancing parameters to be biased toward IN for investigating the virtual \textit{over-style-normalization} situation where our model becomes much more normalized beyond the styles of the source domain. On the contrary, we also lead to the parameters to be biased toward BN for exploring the virtual \textit{under-style-normalization} situation in which our model fails to distinguish identities at the unexpected styles. In this way, we design a complementary objective function that can trigger both challenging cases, which provides an opportunity to escape from the local-minima by overcoming the harsh situations at the unseen-like meta-test domains.

To implement this, we first suggest the losses from the perspective of \textit{over-style-normalization}. Our main point is to enhance intra-domain diversity and disarrange inter-domain distributions like confusing multiple styles. To spread the feature distribution for each domain, we introduce an intra-domain scatter loss as follows:
\vspace{-0.25cm}
\begin{equation}
\fontsize{9.5}{10}\selectfont
\mathcal{L}_\text{scat}(\mathcal{X}_{S}; \theta) = 
\frac{1}{N_{S}}\sum_{k=1}^{K_{S}} {\sum_{i=1}^{N_S^k}}
cos\big(\bm{f}_i^k, \bar{\bm{f}}^k\big),
\vspace{-0.15cm}
\label{eq:scatter}
\end{equation}
\noindent
where $\bar{\bm{f}}^k$ denotes the mean feature vector (centroid) of domain $k$ in a mini-batch, $K_S$ is the number of meta-train domains, $N_S^k$ is the number of meta-train samples for domain $k$, $N_S$ is the number of all meta-train samples, and $cos(\bm{a},\bm{b}) = \bm{a}\cdot\bm{b} / \Vert \bm{a} \Vert \Vert \bm{b} \Vert$.

In addition to this, we propose an inter-domain shuffle loss for supporting the virtual effect of style normalization further. This loss pulls the negative sample of the inter-domain and pushes the negative sample of the intra-domain, so that the inter-domain distributions are shuffled. It is expressed as follows:
\vspace{-0.30cm}
\begin{equation}
\fontsize{9.5}{10}\selectfont
\mathcal{L}_\text{shuf}(\mathcal{X}_{S}; \theta)=
\frac{1}{N_{S}}\sum_{i=1}^{N_{S}}
l_{s}\big(d(\bm{f}_i^a, \bm{f}_i^{n-}) - d(\bm{f}_i^a, \bm{f}_i^{n+})\big), 
\vspace{-0.20cm}
\label{eq:shuffle}
\end{equation}
where $\bm{f}_i^a$, $\bm{f}_i^{n-}$, and $\bm{f}_i^{n+}$ indicate the feature representations of an anchor sample, an inter-domain negative sample, and an intra-domain negative sample. $d(\cdot, \cdot)$ is the Euclidean distance, and $l_{s}$ is the softplus function \cite{dugas2001incorporating}.

At the same time, we add the triplet loss $\mathcal{L}_\text{tr}(\mathcal{X}_{S}; \theta)$ from perspective of \textit{under-style-normalization}. It enhances intra-class compactness regardless style differences. The overall loss for meta-train is as follows:
\vspace{-0.15cm}
\begin{equation}
\fontsize{9.5}{10}\selectfont
\mathcal{L}_\text{mtr} (\mathcal{X}_{S} ; \theta) = \mathcal{L}_\text{scat}(\mathcal{X}_{S}; \theta) + \mathcal{L}_\text{shuf}(\mathcal{X}_{S}; \theta) + \mathcal{L}_\text{tr}(\mathcal{X}_{S}; \theta).
\vspace{-0.20cm}
\label{eq:mtrain}
\end{equation}
The combination of the triplet loss alleviates the excessive discrimination reduction problem caused by $\mathcal{L}_\text{scat}(\mathcal{X}_{S}; \theta)$ and $\mathcal{L}_\text{shuf}(\mathcal{X}_{S}; \theta)$.

Based on the meta-train loss $\mathcal{L}_\text{mtr}$, we inner-update the balancing parameters from $\theta_\rho$ to $\theta'_\rho$ as follows:
\vspace{-0.2cm}
\begin{equation}
\fontsize{9.5}{10}\selectfont
\theta_\rho' = \theta_\rho - \beta \nabla_{\theta_\rho} \mathcal{L}_\text{mtr} (\mathcal{X}_{S} ; \theta_f, \theta_\rho),
\vspace{-0.15cm}
\label{eq:mtrain_update}
\end{equation}
\noindent 
where $\beta$ is a learning rate for the inner-level optimization. 

\textbf{Cyclic inner-updates:}
In the general MAML-based DG methods~\cite{li2017learning, balaji2018metareg}, the learning rate $\beta$ is pre-defined as a constant. To promote the diversity of virtual simulations, we adopt the cyclical learning rate~\cite{smith2017cyclical} for the inner-level optimization, which we call it a cyclical inner-updating method. More specifically, this method affects how much the balancing parameters are updated. Therefore, it diversifies generalization scenarios from the easy case by a small $\beta$ to the difficult case by a large $\beta$.

\textbf{Meta-test:}
After moving the balancing parameters in the inner-level optimization step, we evaluate our model at the unseen-like samples  $\mathcal{X}_T$ from meta-test domains $\mathcal{D}_\text{mte}$. In this step, we can examine various generalization scenarios depending on the movement of the balancing parameters. To validate a retrieval task effectively, we employ the triplet loss with the updated balancing parameters $\theta'_\rho$ as follows: 
\vspace{-0.15cm}
\begin{equation}
\fontsize{9.5}{10}\selectfont
\theta_\rho \leftarrow \theta_\rho - \gamma \nabla_{\theta_\rho} \mathcal{L}_\text{tr} (\mathcal{X}_{T} ; \theta_f, \theta'_\rho).
\vspace{-0.15cm}
\label{eq:mtest_update}
\end{equation}
\noindent
From the equation, we meta-update the balancing parameters to overcome the virtual simulations. Eventually, our model learns to generalize the balancing parameters by the final optimization process. As these meta-learning and base model updating steps are alternately trained, the generalization capability to unseen domains is improved effectively.

\section{Experiments}
\label{sec:Experiments}
\vspace{-0.10cm}

\subsection{Datasets and Settings}
\vspace{-0.10cm}

\textbf{Datasets:}
\label{sec:dataset}
To evaluate the generalization capability of our method, we employ the large-scale domain generalization (DG) Re-ID benchmark~\cite{song2019generalizable} and the cross-domain Re-ID problem. In the large-scale DG Re-ID setting, source datasets contain CUHK02~\cite{li2013locally}, CUHK03~\cite{li2014deepreid}, Market-1501~\cite{zheng2015scalable}, DukeMTMC-ReID~\cite{zheng2017unlabeled}, and CUHK-SYSU PersonSearch~\cite{xiao2016end}, and target datasets include VIPeR~\cite{gray2008viewpoint}, PRID~\cite{hirzer2011person}, GRID~\cite{loy2009multi}, and QMUL i-LIDS~\cite{wei2009associating}. All images in the source datasets are used for training regardless of train/test splits, and the total number of identities is $M = 18,\!530$ with $N = 121,\!765$ images. In the cross-domain Re-ID setting, we employ Market-1501~\cite{zheng2015scalable} and DukeMTMC-ReID~\cite{zheng2017unlabeled}. We alternately construct the two datasets into source and target domains. At this time, we regard each camera within a dataset as an individual domain.

\begin{table*}[tb]
	\caption{Performance (\%) comparison with the state-of-the-arts on the large-scale DG Re-ID benchmark, where `$\dagger$' is based on ResNet-50.}
	\vspace{0.10cm}
	\centering
	\resizebox{\textwidth}{!}{
		\begin{tabular}{c||cc||cccc||cccc||cccc||cccc}
			 \Xhline{3\arrayrulewidth}
			 \multirow{3}{*}{Method} & \multicolumn{18}{c}{Large-scale domain generalization Re-ID (multi-source DG)} \\ \cline{2-19} 
			 & \multicolumn{2}{c||}{Average}  & \multicolumn{4}{c||}{Target: VIPeR (V)~\cite{gray2008viewpoint}} & \multicolumn{4}{c||}{Target: PRID (P)~\cite{hirzer2011person}} & \multicolumn{4}{c||}{Target: GRID (G)~\cite{loy2009multi}} & \multicolumn{4}{c}{Target: i-LIDS (I)~\cite{wei2009associating}} \\ \cline{2-19} 
			 & R-1 & \emph{m}AP & R-1 & R-5 & R-10 & \emph{m}AP & R-1 & R-5 & R-10 & \emph{m}AP & R-1 & R-5 & R-10 & \emph{m}AP & R-1 & R-5 & R-10 & \emph{m}AP \\ \hline

DIMN~\cite{song2019generalizable}       	            & 47.5 & 57.9 & 51.2 & 70.2 & 76.0 & 60.1 & 39.2 & 67.0 & 76.7 & 52.0 & 29.3 & 53.3 & 65.8 & 41.1 & 70.2 & 89.7 & 94.5 & 78.4 \\
AugMining~\cite{tamura2019augmented}                    & 51.8 & -    & 49.8 & 70.8 & 77.0 &  -   & 34.3 & 56.2 & 65.7 & -    & 46.6 & {\textbf{67.5}} & {{76.1}} & -    & 76.3 & {{93.0}} & {{95.3}} & -    \\
Switchable (BN+IN)~\cite{luo2019switchable}   		    & 57.0 & {{65.6}} & 51.6 & {{72.9}} & {{80.8}} & {{61.4}} & 59.6 & {{78.6}} & {{90.1}} & {{69.4}} & 39.3 & 58.8 & 68.1 & 48.1 & 77.3 & 91.2 & 94.8 & {{83.5}} \\ 
DualNorm~\cite{jia2019frustratingly}   			        & 57.6 & 61.8 & 53.9 & 62.5 & 75.3 & 58.0 & 60.4 & 73.6 & 84.8 & 64.9 & 41.4 & 47.4 & 64.7 & 45.7 & 74.8 & 82.0 & 91.5 & 78.5 \\ 
DDAN~\cite{chen2020dual}                    	        & 59.0 & 63.1 & 52.3 & 60.6 & 71.8 & 56.4 & 54.5 & 62.7 & 74.9 & 58.9 & {\textbf{50.6}} & 62.1 & 73.8 & {{55.7}} & {{78.5}} & 85.3 & 92.5 & 81.5 \\
DDAN~\cite{chen2020dual} w/ \cite{jia2019frustratingly} & {{60.9}} & 65.1 & {{56.5}} & 65.6 & 76.3 & 60.8 & {{62.9}} & 74.2 & 85.3 & 67.5 & 46.2 & 55.4 & 68.0 & 50.9 & 78.0 & 85.7 & 93.2 & 81.2 \\ 
\textbf{MetaBIN (Ours)}                                          & {\textbf{64.7}} & {\textbf{72.3}} & {\textbf{56.9}} & {\textbf{76.7}} & {\textbf{82.0}} & {\textbf{66.0}} & {\textbf{72.5}} & {\textbf{88.2}} & {\textbf{91.3}} & {\textbf{79.8}} & {{49.7}} & {\textbf{67.5}} & {\textbf{76.8}} & {\textbf{58.1}} & {\textbf{79.7}} & {\textbf{93.3}} & {\textbf{97.3}} & {\textbf{85.5}} \\ \hline
SNR${}^\dagger$~\cite{jin2020style}                     & 57.3 & {{66.4}} & 52.9 & -    & -    & {{61.3}} & 52.1 & -    & -    & {{66.5}} & 40.2 & -    & -    & {{47.7}} & {\textbf{84.1}} & -    & -    & {\textbf{89.9}} \\
DualNorm${}^\dagger$~\cite{jia2019frustratingly}   		& {{62.7}} & -    & {{59.4}} & -    & -    & -    & {{69.6}} & -    & -    & -    & {{43.7}} & -    & -    & -    & 78.2 & -    & -    & -    \\ 
\textbf{MetaBIN${}^\dagger$ (Ours)}                              & {\textbf{66.0}} & {\textbf{73.6}} & {\textbf{59.9}} & {\textbf{78.4}} & {\textbf{82.8}} & {\textbf{68.6}} & {\textbf{74.2}} & {\textbf{89.7}} & {\textbf{92.2}} & {\textbf{81.0}} & {\textbf{48.4}} & {\textbf{70.3}} & {\textbf{77.2}} & {\textbf{57.9}} & {{81.3}} & {\textbf{95.0}} & {\textbf{97.0}} & {{87.0}} \\ \Xhline{3\arrayrulewidth}
		\end{tabular}
	}
 \vspace{-0.15cm}
	\label{table:result}
\end{table*}

\textbf{Implementation details:}
\label{sec:implementation}
For a fair comparison, we adopted MobileNetV2~\cite{sandler2018mobilenetv2} with a multiplier of 1.4 and ResNet-50~\cite{he2016deep} as backbone networks. The weights are pre-trained on ImageNet~\cite{deng2009imagenet}. All balancing parameters are initialized to 1. For training, each image is resized to 256 $\times$ 128. We select 16 samples for each domain in a mini-batch. In the base model updating step, all images in a mini-batch are used. However, in a meta-learning pipeline, domains are separated into meta-train and meta-test from another mini-batch. The label-smoothing parameter is 0.1 and the margin in the triplet loss is 0.3. For updating our base model, we use the SGD optimizer with a momentum of 0.9 and a weight decay of $5 \times 10^{-4}$. Its initial learning rate $\alpha$ is 0.01, which is warmed up for 10 epochs as \cite{luo2019bag} and decayed to its 0.1$\times$ and 0.01$\times$ at 40 and 70 epochs. In the meta-learning step, the balancing parameters are updated by another SGD optimizer without momentum and weight decay. The meta-train step-size $\beta$ oscillates back and forth in the range [0.001, 0.1] with the triangular policy~\cite{smith2017cyclical}. The meta-test step-size $\gamma$ is fixed to 0.1. The training stops at 100 epochs. To speed up the training process and increase memory efficiency, we use the automatic mixed-precision training~\cite{micikevicius2017mixed} in the entire process and the first-order approximations~\cite{finn2017model} in meta-optimization. All experiments are conducted on an NVIDIA Titan Xp GPU using Pytorch.

\textbf{Evaluation metrics:}
We follow the common evaluation metrics for Re-ID as mean Average Precision (mAP) and Cumulative Matching Characteristic (CMC) at Rank-$k$.

\subsection{Comparison with State-of-the-art Methods}
\label{sec:sota}
\vspace{-0.05cm}

\textbf{Large-scale DG Re-ID benchmark:} We evaluate our MetaBIN framework on the large-scale domain generalization Re-ID benchmark~\cite{song2019generalizable}, which is shown in Table~\ref{table:result}. For a fair comparison, we provide both results based on MobileNetV2 and ResNet-50. All Re-ID models have been evaluated individually on each target dataset. Among the competitors, DualNorm~\cite{jia2019frustratingly} achieved relatively good performance on VIPeR and PRID, and DDAN~\cite{chen2020dual} also obtained comparable performance on GRID. In addition, SNR~\cite{jin2020style} achieved impressive results on the i-LIDS dataset in the experiment based on ResNet-50. Nevertheless, our MetaBIN method outperformed all competing methods by a significant margin in the average performance. The main reasons come from two aspects: 1) Our MetaBIN framework, which simulates the virtual scenarios in a meta-learning pipeline, effectively enhances the generalization capability to novel domains; 2) With the addition of simple balancing parameters and the separation of learning episodes, we can solve the challenging heterogeneous DG problem without synthetic data augmentation or complicated network design.

\begin{table}[tb]
	\caption{Performance (\%) comparison with the state-of-the-arts on the cross-domain Re-ID problem.}
	\vspace{0.10cm}
	\centering
	\resizebox{0.47\textwidth}{!}{
		\begin{tabular}{c||cccc||cccc}
			 \Xhline{3\arrayrulewidth}
			\multirow{3}{*}{Method} & \multicolumn{8}{c}{Cross-domain Re-ID (single-source DG)} \\ \cline{2-9} 
			 & \multicolumn{4}{c||}{Market1501 $\rightarrow$ DukeMTMC} & \multicolumn{4}{c}{DukeMTMC $\rightarrow$ Market1501} \\ \cline{2-9} 
			 & R-1 & R-5 & R-10 & \emph{m}AP & R-1 & R-5 & R-10 & \emph{m}AP \\ \hline
IBN-Net~\cite{pan2018two}                       & 43.7 & 59.1 & 65.2 & 24.3 & 50.7 & 69.1 & 76.3 & 23.5  \\ 
OSNet~\cite{zhou2019learning}                   & 44.7 & 59.6 & 65.4 & 25.9 & 52.2 & 67.5 & 74.7 & 24.0  \\ 
OSNet-IBN~\cite{zhou2019learning}               & 47.9 & 62.7 & 68.2 & 27.6 & 57.8 & 74.0 & 79.5 & 27.4  \\ 
CrossGrad~\cite{shankar2018generalizing}        & 48.5 & 63.5 & 69.5 & 27.1 & 56.7 & 73.5 & 79.5 & 26.3  \\ 
QAConv~\cite{liao2019interpretable}             & 48.8 & -    & -    & 28.7 & 58.6 & -    & -    & 27.2  \\ 
L2A-OT~\cite{zhou2020learning}                  & 50.1 & 64.5 & 70.1 & 29.2 & 63.8 & 80.2 & 84.6 & 30.2  \\ 
OSNet-AIN~\cite{zhou2019learning}               & 52.4 & 66.1 & 71.2 & 30.5 & 61.0 & 77.0 & 82.5 & 30.6  \\ 
SNR~\cite{jin2020style}                         & 55.1 & -    & -    & \textbf{33.6} & 66.7 & -    & -    & 33.9  \\ 
\textbf{MetaBIN (Ours)}                                  & \textbf{55.2} & \textbf{69.0} & \textbf{74.4} & 33.1 & \textbf{69.2} & \textbf{83.1} & \textbf{87.8} & \textbf{35.9} \\ 

\Xhline{3\arrayrulewidth}
		\end{tabular}
	}
 \vspace{-0.15cm}
	\label{table:single-resnet}
\end{table}

\textbf{Cross-domain Re-ID problem:} To demonstrate the superiority of our method, we additionally compare our MetaBIN framework with a variety of state-of-the-art methods in the cross-domain Re-ID problem, which is shown in Table~\ref{table:single-resnet}. `Market1501 $\rightarrow$ DukeMTMC' indicates that Market-1501~\cite{zheng2015scalable} is a labeled source domain and DukeMTMC-ReID~\cite{zheng2017unlabeled} is an unseen target domain. 
In other words, we need to generalize a Re-ID model so that it works well on the unseen target dataset using only the images on a single source dataset. The setting of cross-domain Re-ID is somewhat different from that of large-scale domain generalization. First, the style variation within a single dataset is relatively small. Therefore, we increased the meta-optimization step size to 0.2 on Market-1501~\cite{zheng2015scalable} and 0.5 on DukeMTMC-ReID~\cite{zheng2017unlabeled}. Second, some identities are simultaneously distributed on different camera domains. Thus, we added the cross-entropy loss to the meta-train and meta-test losses for enhancing the ID-discrimination ability. Third, since the number of domains is different from the previous one, we divided the entire camera domains in half for each dataset. Finally, for a fair comparison with SNR~\cite{jin2020style}, we employed ResNet-50~\cite{he2016deep} with color jittering. Table~\ref{table:single-resnet} show that our MetaBIN framework achieved comparable performance on both settings. In the end, we prove that our framework has the potential to improve the generalization capability even on the single-source dataset with relatively small domain discrepancy.

\subsection{Ablation study}
\label{sec:ablation}
\vspace{-0.10cm}

We perform comprehensive ablation studies to show the effectiveness of our MetaBIN framework and detailed components through the average performance on the large-scale DG Re-ID benchmark.

\textbf{Influence of model components:}
Table \ref{table:ablation} reports the ablation results of our novel framework including meta-train losses and the cyclic inner-updating method. We employed the cross-entropy and triplet losses as a baseline. We observed the following aspects: 1) Our novel MetaBIN framework that updates balancing parameters in meta-learning improves the generalization capability with a large performance gap by alleviating the overfitting problem at given source domains; 2) In simulating real train-test domain shifts, the triplet loss is more suitable than the cross-entropy loss for DG Re-ID, which is different from the general homogeneous DG methods~\cite{li2017learning, balaji2018metareg}; 3) Our meta-train losses are complementary to each other and have a synergistic effect with the cyclic inner-updating method.

\begin{table}[tb]
	\centering
	\caption{Ablation studies of our MetaBIN framework in the average performance on the large-scale DG Re-ID benchmark.}
	\vspace{0.10cm}
	\resizebox{.42\textwidth}{!}{
	\begin{tabular}{c||c|c|c||cc} \Xhline{3\arrayrulewidth}
	Method & $\mathcal{L}_\text{mtr}$ & $\mathcal{L}_\text{mte}$ & $\beta$ & R-1 & \emph{m}AP \\ \hline
     BN & - & - & - & 50.9 & 59.5 \\
     MetaBIN & $\mathcal{L}_\text{ce}$                          & $\mathcal{L}_\text{ce}$                       & fixed                 & 60.6      & 69.4 \\
     MetaBIN & $\mathcal{L}_\text{ce},\mathcal{L}_\text{tr}$    & $\mathcal{L}_\text{ce},\mathcal{L}_\text{tr}$ & fixed                 & 62.0      & 69.9 \\
     MetaBIN & $\mathcal{L}_\text{tr}$                          & $\mathcal{L}_\text{tr}$                       & fixed                 & 62.8      & 70.8 \\
     MetaBIN & $\mathcal{L}_\text{tr},\mathcal{L}_\text{scat}$ & $\mathcal{L}_\text{tr}$ & fixed                & 63.0      & 71.0 \\ 
     MetaBIN & $\mathcal{L}_\text{tr},\mathcal{L}_\text{shuf}$ & $\mathcal{L}_\text{tr}$ & fixed                & 63.1      & 71.0 \\ 
     MetaBIN & $\mathcal{L}_\text{tr},\mathcal{L}_\text{scat},\mathcal{L}_\text{shuf}$ & $\mathcal{L}_\text{tr}$ & fixed                & 63.5      & 71.3 \\ 
     \textbf{MetaBIN} & \textbf{$\mathcal{L}_\text{tr},\mathcal{L}_\text{scat},\mathcal{L}_\text{shuf}$} & \textbf{$\mathcal{L}_\text{tr}$} & \textbf{cyclic}  & \textbf{64.7}       & \textbf{72.3} \\ \Xhline{3\arrayrulewidth}
	\end{tabular}}
	\label{table:ablation}
\vspace{-0.1cm}
\end{table}

\begin{figure}[t]
\begin{center}
\includegraphics[width=1.0\linewidth]{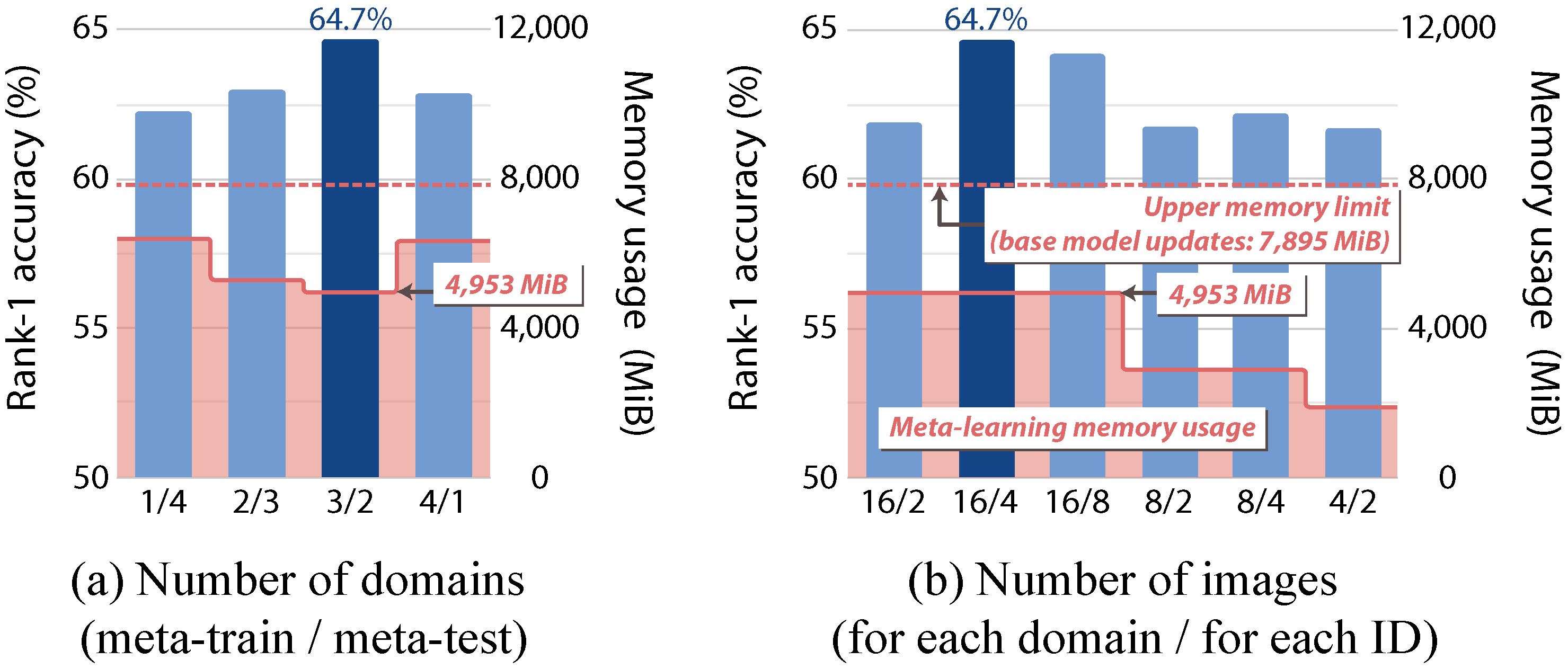}
\end{center}
\vspace{-0.3cm}
\caption{Performance (\%) and memory usage (MiB) analysis according to the sampling manner of domains and images.}
\vspace{-0.3cm}
\label{fig:domain}
\end{figure}

\textbf{Influence of domain-level sampling:}
In this part, we compare the differences depending on the sampling manner of domains and images in the meta-learning step. In Fig. \ref{fig:domain} (a), whereas the simulation quality deteriorates if the number of meta-train domains is too small, the generalization capability to overcome the overfitting issues becomes insufficient if the number of meta-test domains is too small. Therefore, the domain sampling between meta-train and meta-test should be adequately balanced. Our method achieved the highest performance and the lowest memory usage when the number of meta-train and meta-test domains is 3 and 2, respectively. Figure \ref{fig:domain} (b) shows the performance and memory usage according to the number of images per domain in a mini-batch. It is noteworthy that we chose the highest performance case under the condition that does not exceed the memory usage (7,895MiB) of updating a base model. In other words, the proposed meta-learning pipeline does not increase the memory usage, since the meta-learning and base model updating steps are divided. Thus, it has an advantage in terms of memory consumption over the MLDG method~\cite{li2017learning} of meta-updating the entire parameters.

\begin{table}[tb]
	\caption{\centering{Performance (\%) comparison in a meta-learning pipeline.}}
	\vspace{0.10cm}
	\centering
	\resizebox{.39\textwidth}{!}{
	\begin{tabular}{c|c|c|c||cc} \Xhline{3\arrayrulewidth}
	Method & $\mathcal{L}_\text{base}$ & MLDG~\cite{li2017learning} & cyclic $\beta$ & R-1 & \emph{m}AP \\ \hline
     \multirow{6}{*}{BN} & $\mathcal{L}_\text{ce}$  & \xmark & \xmark  & 50.2         & 59.6  \\
      & $\mathcal{L}_\text{ce}$  & \cmark & \xmark  & 50.5         & 59.2 \\
      & $\mathcal{L}_\text{ce}$ & \cmark & \cmark   & 52.3         & 60.9 \\ \cline{2-6}
      & $\mathcal{L}_\text{ce}, \mathcal{L}_\text{tr}$ & \xmark & \xmark  & 50.9         & 59.5  \\
      & $\mathcal{L}_\text{ce}, \mathcal{L}_\text{tr}$  & \cmark & \xmark & 52.2         & 61.2 \\
      & $\mathcal{L}_\text{ce}, \mathcal{L}_\text{tr}$  & \cmark & \cmark  & 53.6         & 61.8 \\  \hline
     \multirow{3}{*}{BIN~\cite{nam2018batch}} & $\mathcal{L}_\text{ce}, \mathcal{L}_\text{tr}$  & \xmark & \xmark                                         & 54.8         & 63.1 \\ 
     & $\mathcal{L}_\text{ce}, \mathcal{L}_\text{tr}$ & \cmark & \xmark                 & 57.9         & 65.7 \\ 
     & $\mathcal{L}_\text{ce}, \mathcal{L}_\text{tr}$ & \cmark & \cmark  & 58.4           & 66.3 \\ \hline
     \multicolumn{4}{c||}{MetaBIN (replace with BIN~\cite{nam2018batch}) }            & 60.6         & 68.8 \\ 
     \multicolumn{4}{c||}{MetaBIN (w/o episode separation)}            & 60.9         & 69.1 \\ \hline
     \multicolumn{4}{c||}{\textbf{MetaBIN}}    & \textbf{64.7}         & \textbf{72.3} \\ \Xhline{3\arrayrulewidth}
	\end{tabular}}
	\label{table:MAML}
	\vspace{-0.25cm}
\end{table}

\begin{figure*}[t]
\begin{center}
\includegraphics[width=1.0\linewidth]{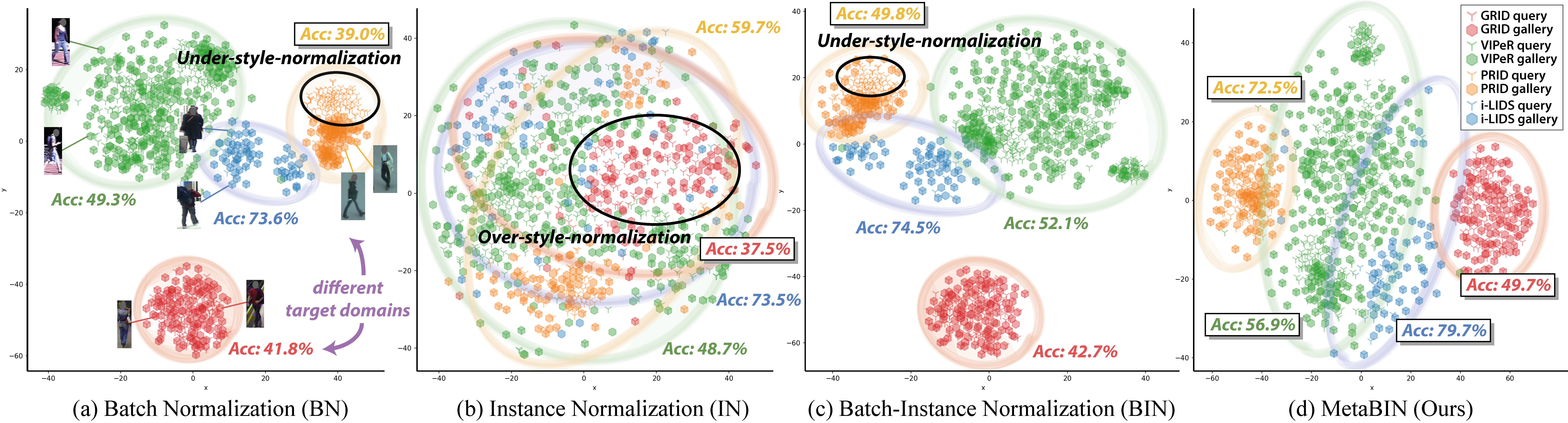}
\end{center}
\vspace{-0.35cm}
\caption{The t-SNE visualization of the embedding vectors on four unseen target datasets (VIPeR, PRID, GRID, and i-LIDS). Trained models are evaluated individually for each dataset, and query and gallery samples are expressed in different shapes. Best viewed in color.}
\vspace{-0.48cm}
\label{fig:tsne}
\end{figure*}

\begin{table}[tb]
	\caption{Performance (\%) comparison with normalization methods in DG and supervised settings, where `S' is single normalization, `N' is non-parametric normalization, `P' is parametric normalization, `BN+IN half' is a channel-wise combination of BN and IN.}
	\vspace{0.10cm}
	\centering
	\resizebox{.46\textwidth}{!}{
	\begin{tabular}{c|c||p{0.75cm}p{0.75cm}||p{1.6cm}p{1.6cm}} 
	\Xhline{3\arrayrulewidth}
	\multicolumn{2}{c||}{\multirow{2}{*}{Method}} & \multicolumn{2}{c||}{Large-scale DG} & \multicolumn{2}{c}{Supervised (Market1501)} \\ \cline{3-6} 
	 \multicolumn{2}{c||}{}                                            &\hfil R-1  &\hfil \emph{m}AP   &\hfil R-1      &\hfil \emph{m}AP \\ \hline
     \multirow{2}{*}{S} & BN                                                     &\hfil 50.9 &\hfil 59.5            &\hfil 87.2     &\hfil 67.9 \\
     & IN                                                     &\hfil 54.9 &\hfil 63.3            &\hfil 71.9     &\hfil 46.1 \\  \hline
     \multirow{2}{*}{N} & DualNorm~\cite{jia2019frustratingly}                   &\hfil 57.6 &\hfil 61.8            &\hfil 82.6     &\hfil 57.2 \\
     & BN+IN half                                             &\hfil 56.5 &\hfil 65.3            &\hfil 79.5     &\hfil 53.9 \\  \hline
     \multirow{2}{*}{P} & BIN~\cite{nam2018batch}                    &\hfil 54.8 &\hfil 63.1            &\hfil 87.5     &\hfil 67.8 \\
     & \textbf{MetaBIN (Ours)}                                          &\hfil \textbf{64.7} &\hfil \textbf{72.3}           &\hfil \textbf{87.9}      &\hfil \textbf{68.5} \\ \Xhline{3\arrayrulewidth}
	\end{tabular}}
	\label{table:norm}
	\vspace{-0.30cm}
\end{table}

\subsection{Further Analysis}
\vspace{-0.10cm}

\textbf{Analysis from the perspective of MAML:} Table \ref{table:MAML} presents the experimental results in a meta-learning pipeline. MLDG~\cite{li2017learning} is a representative MAML-based method designed for homogeneous DG. We observed that MLDG does not improve performance much, which means that it is difficult to directly apply the conventional homogeneous DG method to the challenging DG Re-ID task. On the other hand, when the meta-learning approach is combined with the triplet loss or a batch-instance normalization layer, its performance has been meaningfully improved. Thus, these methods are valid for solving the DG Re-ID problem. However, we note that the generalization potential of the existing BIN model~\cite{nam2018batch} was insufficient compared to that of our model. The proposed cyclic inner-updating method has made significant improvements in all cases, indicating an effective model-free method. Finally, the separation of training episodes contributed to improving performance considerably. It demonstrates that selectively updating only the balancing parameters in the complicated meta-learning pipeline brings training stability and improves the generalization capacity.

\begin{figure}[t]
\begin{center}
\includegraphics[width=1.0\linewidth]{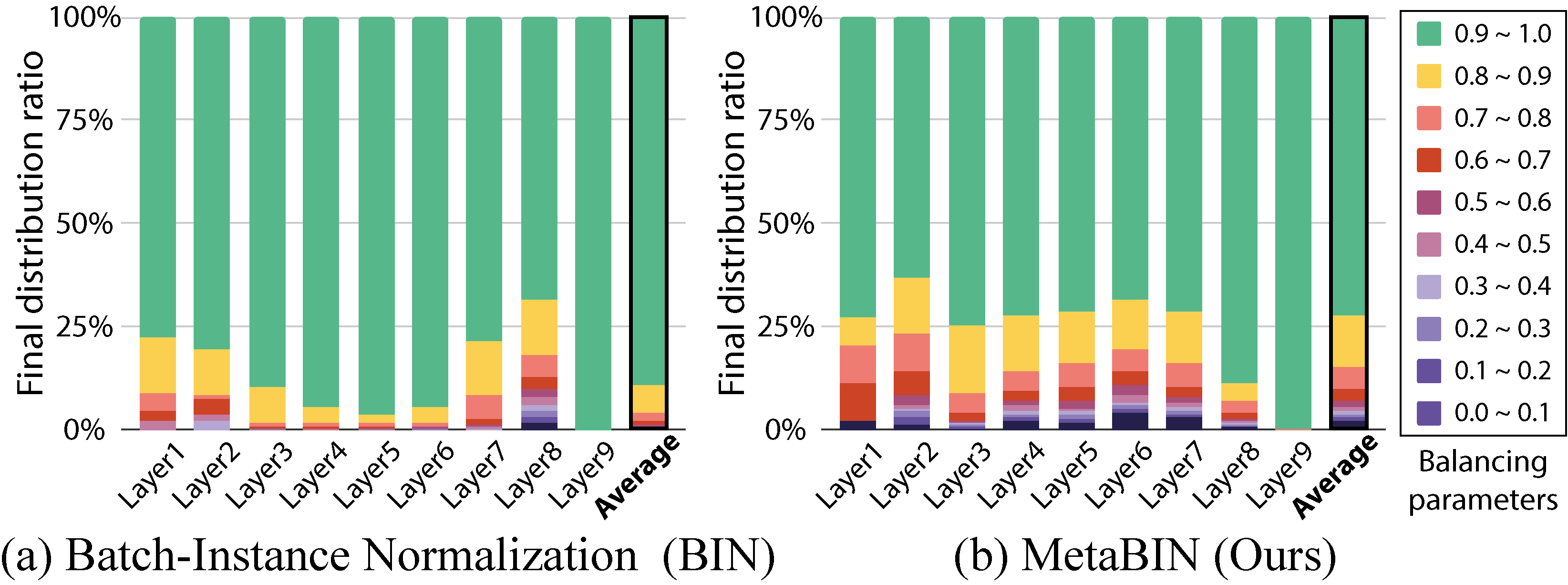}
\end{center}
\vspace{-0.35cm}
\caption{Analysis on the final distribution ratios of balancing parameters for each layer. Best viewed in color.}
\vspace{-0.38cm}
\label{fig:BIN}
\end{figure}

\textbf{Analysis from the perspective of normalization:} 
We analyze our MetaBIN framework and other normalization methods through the performance and t-SNE visualization~\cite{maaten2008visualizing}. First, we compare BN with IN as single normalization techniques. We observed \textit{under-style-normalization} in Fig.~\ref{fig:tsne}~(a) and \textit{over-style-normalization} in Fig.~\ref{fig:tsne}~(b), and it turned out that the performance plummeted when these scenarios occurred as reported in Table~\ref{table:norm}. Furthermore, since IN filtered out discriminative information too much, its performance had been greatly degraded in the supervised setting. Meanwhile, the non-parametric or parametric combination of BN and IN alleviated these problems. The non-parametric methods in the DG experiment and the parametric methods in the supervised setting showed relatively high performance. However, a poorly generalized case was also observed at the BIN method~\cite{nam2018batch}, as expressed in Fig.~\ref{fig:tsne}~(c). On the other hand, our MetaBIN method overcame the overfitting issue by simulating unsuccessful generalization in the meta-learning pipeline, eventually surpassing all normalization methods in the DG experiment. Figure~\ref{fig:tsne}~(d) shows that our method has shorter query-gallery distances than those of other methods. Besides, our method improves performance even in the supervised setting. Thus, we demonstrate that our MetaBIN method is generalizable and practical for real-world situations.

\textbf{Analysis on the balancing parameters:}
To understand how normalization layers are generalized, we investigate the final distribution ratio of balancing parameters, as shown in Fig. \ref{fig:BIN}. Both experiments have been conducted based on MobileNetV2~\cite{sandler2018mobilenetv2} trained on the large-scale DG Re-ID dataset. Note that all balancing parameters have been initialized to 1 (toward BN). We observed that most of the parameters tend to maintain the BN properties, but only some parameters are biased toward IN. Since the performance improved through this process, we demonstrate that removing useless instance-specific style information contributes to improving generalization capability. In particular, the distribution difference between BIN~\cite{nam2018batch} and MetaBIN shows that deflecting some channels toward IN in the middle layer as well as the shallow layer alleviates the overfitting issue and promotes the generalization capability.

\begin{figure}[t]
\begin{center}
\includegraphics[width=0.84\linewidth]{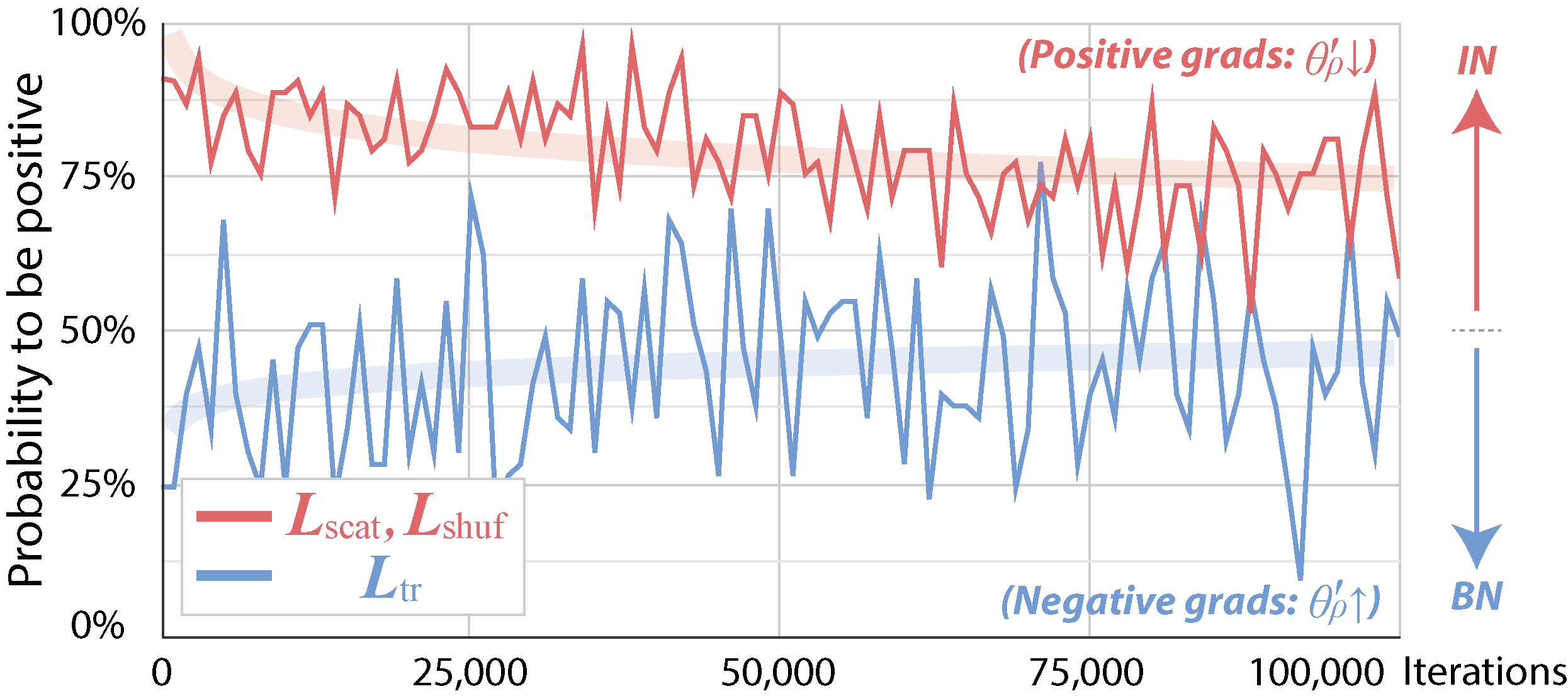}
\end{center}
\vspace{-0.40cm}
\caption{Analysis on the gradients in the inner-optimization step according to the meta-train loss ($\mathcal{L}_\text{tr},\mathcal{L}_\text{scat},\mathcal{L}_\text{shuf}$).}
\vspace{-0.35cm}
\label{fig:mtrain}
\end{figure}

\textbf{Analysis on the meta-train loss:} 
We analyze the virtual simulation scenarios through visualization of gradients calculated in the inner-optimization step, as illustrated in Fig. \ref{fig:mtrain}. First, the intra-domain scatter loss $\mathcal{L}_\text{scat}$ and inter-domain shuffle loss $\mathcal{L}_\text{shuf}$ generate positive gradients, which act to move the balancing parameters in the IN direction. In addition, the triplet loss $\mathcal{L}_\text{tr}$ promotes negative gradients, which cause the parameters to shift in the BN direction. Eventually, we prove that our meta-train losses induce the balancing parameters to the specific directions, which means that unsuccessful generalization scenarios can be deliberately simulated. We also emphasize that combining all these losses helps improve performance, as shown in Table \ref{table:ablation}.

\vspace{-0.10cm}
\section{Conclusion}
\label{Conclusion}
\vspace{-0.15cm}

In this work, we have proposed a novel generalizable person re-identification framework, called Meta Batch-Instance Normalization (MetaBIN). Compared to previous studies, the proposed method alleviates the overfitting problem by investigating unsuccessful generalization scenarios based on our observation. Furthermore, our novel meta-train loss and cyclic inner-updating method diversify virtual simulations and eventually boost generalization capability to unseen domains. Extensive experiments and comprehensive analysis in the single- and multi-source DG settings demonstrate its superiority over state-of-the-art methods.

{\small
\bibliographystyle{ieee_fullname}
\bibliography{egbib}
}

\clearpage

\appendix
  \renewcommand\thesection{\arabic{section}}

\noindent{\LARGE \textbf{Appendix}}
\vspace{5mm}

\vspace{0.8cm}

\section{More Analysis on Hyperparameters}
\vspace{0.3cm}

\begin{figure}[t]
\begin{center}
\vspace{3.0cm}
\includegraphics[width=1.0\linewidth]{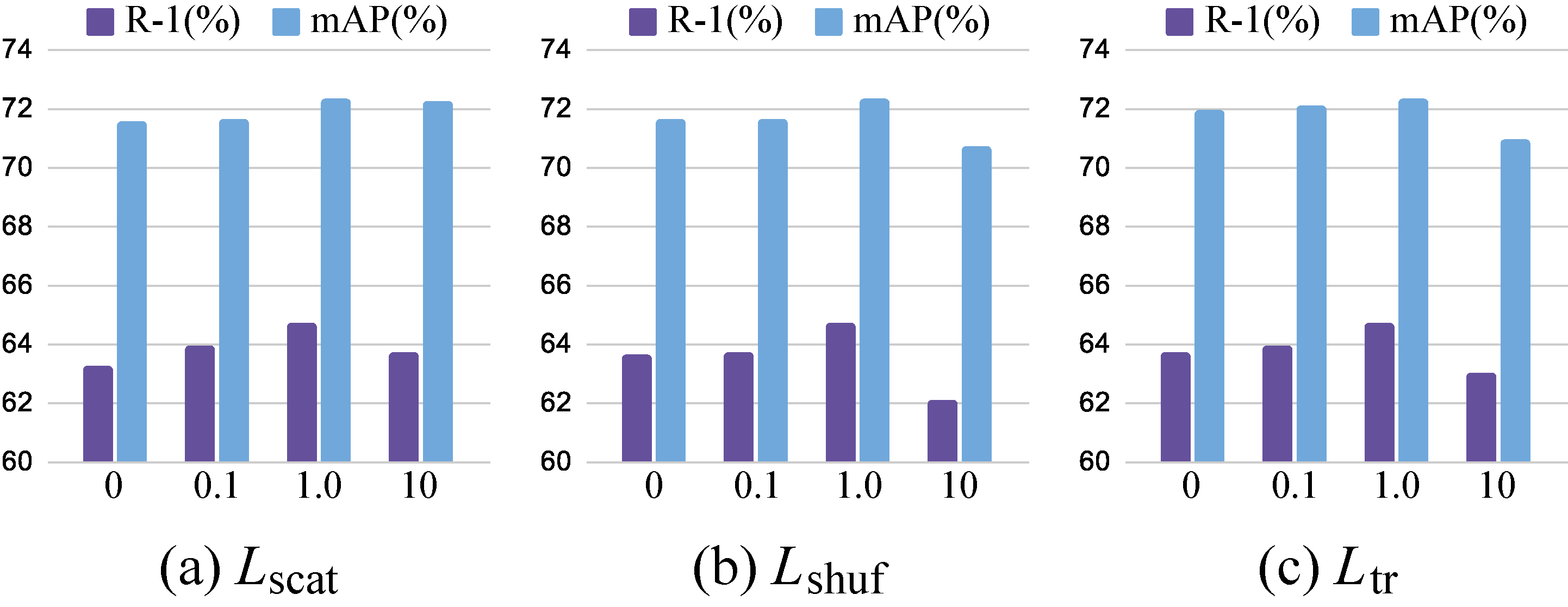}
\end{center}
\vspace{-0.1cm}
\caption{
Performance comparison according to the change of the weight parameters in the meta-train loss. We adjust each weight parameter while fixing the other two weight parameters to $1.0$. 
}
\vspace{0.1cm}
\label{fig:weight}
\end{figure}

\begin{figure}[t]
\begin{center}
\vspace{0.00cm}
\includegraphics[width=1.0\linewidth]{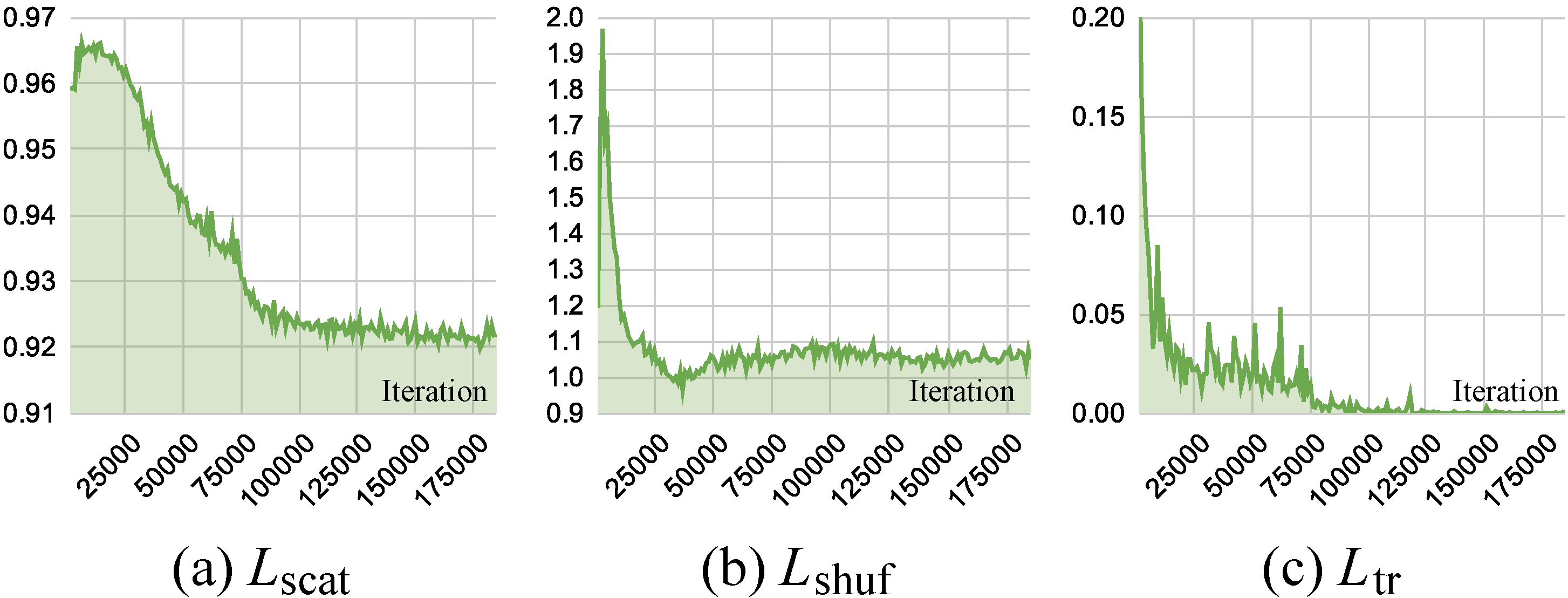}
\end{center}
\vspace{-0.1cm}
\caption{\centering{
The meta-train losses during the training process.
}}
\vspace{0.1cm}
\label{fig:training_loss}
\end{figure}

\textbf{Meta-train loss:} 
The total meta-train loss can be reformulated as $\mathcal{L}_\text{mtr} = \lambda_\text{scat} \mathcal{L}_\text{scat} +  \lambda_\text{shuf} \mathcal{L}_\text{shuf} +  \lambda_\text{tr} \mathcal{L}_\text{tr}$. We compare the performance by changing each of these weights, as expressed in Fig.~\ref{fig:weight}. We observed the highest performance when each weight parameter was $1.0$. In addition, when we assign each weight parameter to $0$ (\ie remove the corresponding loss), its performance deteriorates, which proves that all loss components are essential. Note that when the weights for the inter-domain shuffle loss $\mathcal{L}_\text{shuf}$ and the triplet loss $\mathcal{L}_\text{tr}$ increase, their performances can be even lower. Thus, it is important to balance the weight parameters. For a more detailed analysis, we investigate the changes in the meta-train losses during the training process, as described in Fig. \ref{fig:training_loss}. At the beginning of training, whereas the intra-domain scatter loss $\mathcal{L}_\text{scat}$ and the inter-domain shuffle loss $\mathcal{L}_\text{shuf}$ increase rapidly, the triplet loss $\mathcal{L}_\text{tr}$ decreases dramatically. It shows that the model initially focuses on improving discrimination power. Unlike in the early stage of training, we observed a tendency for all three types of losses to decrease simultaneously. It means that our model becomes generalized as learning progresses.

\vspace{0.3cm}

\begin{figure}[t]
\begin{center}
\vspace{0.00cm}
\includegraphics[width=0.9\linewidth]{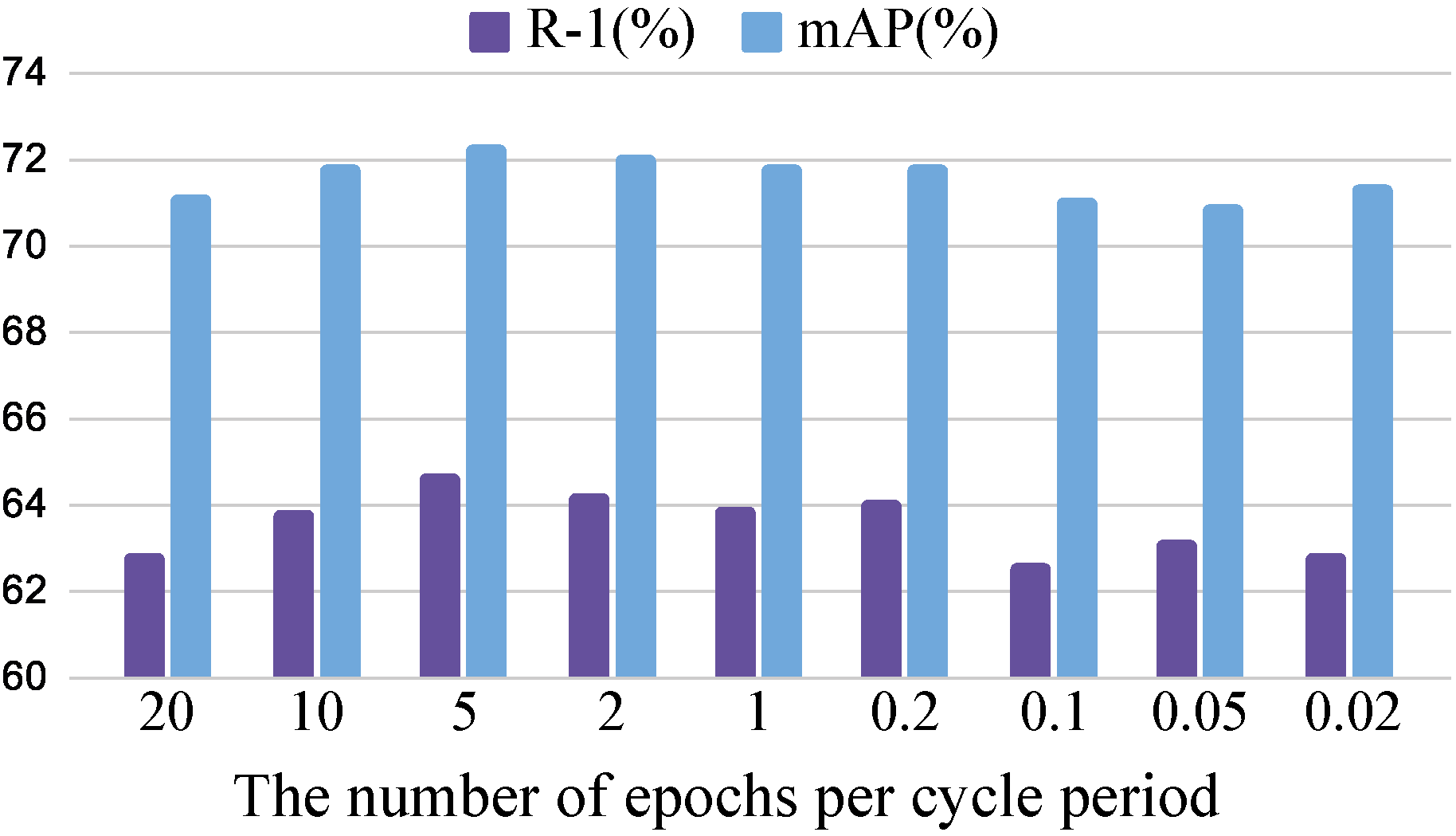}
\end{center}
\vspace{-0.1cm}
\caption{\centering{
Analysis of a cycle period in inner-level optimization.
}}
\vspace{0.1cm}
\label{fig:period_bar}
\end{figure}

\textbf{Cyclic inner-updates:} We applied our cyclic inner-updating method to diversify virtual simulations. Figure \ref{fig:period_bar} shows the performance differences over the cycle period. We observed the highest performance in the case of five epochs, so we set the cycle period to five epochs (\ie $9,\!245$ iterations). In other words, the meta-train step size $\beta$ oscillates back and forth at every five epochs.

\vspace{0.3cm}

\begin{figure*}[t]
\begin{center}
\vspace{0.00cm}
\includegraphics[width=0.96\linewidth]{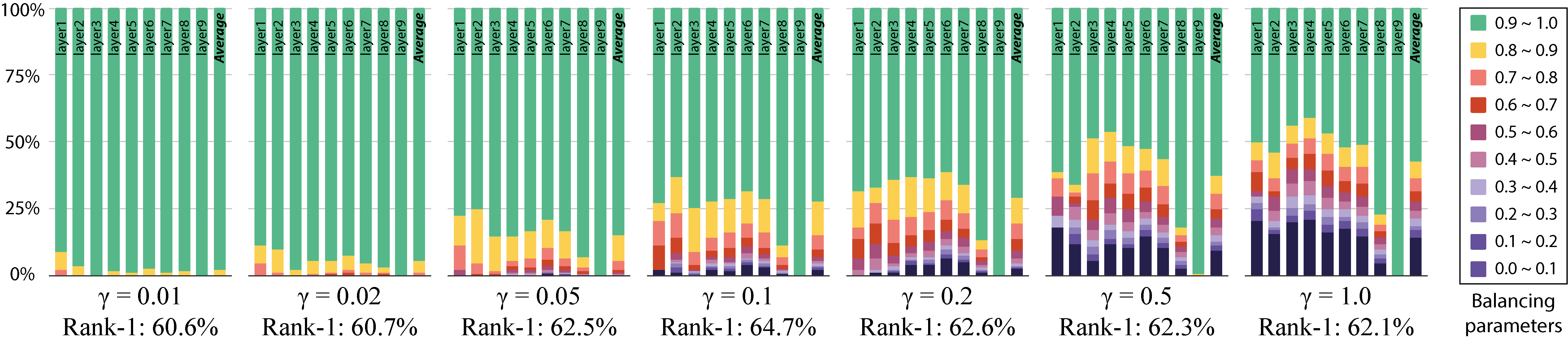}
\end{center}
\vspace{-0.30cm}
\caption{
Analysis on the final distribution ratios of balancing parameters depending on the step size $\gamma$ in meta-optimization.
}
\vspace{-0.10cm}
\label{fig:bin_hist_gamma}
\end{figure*}

\textbf{Step size in meta-optimization:} While the step size $\beta$ of inner-level optimization oscillates, the step size $\gamma$ of final meta-optimization is assigned a fixed value. Figure \ref{fig:bin_hist_gamma} illustrates the final distribution ratios of balancing parameters according to the size of $\gamma$. As the step size increases, the balancing parameters were biased to the instance normalization. Thus, it is important to select an appropriate hyperparameter value considering the style variation between domains. We selected the step size $\gamma$ as $0.1$ and achieved the highest performance with it.

\vspace{3.0cm}

\begin{figure*}[t]
\begin{center}
\vspace{0.00cm}
\includegraphics[width=0.87\linewidth]{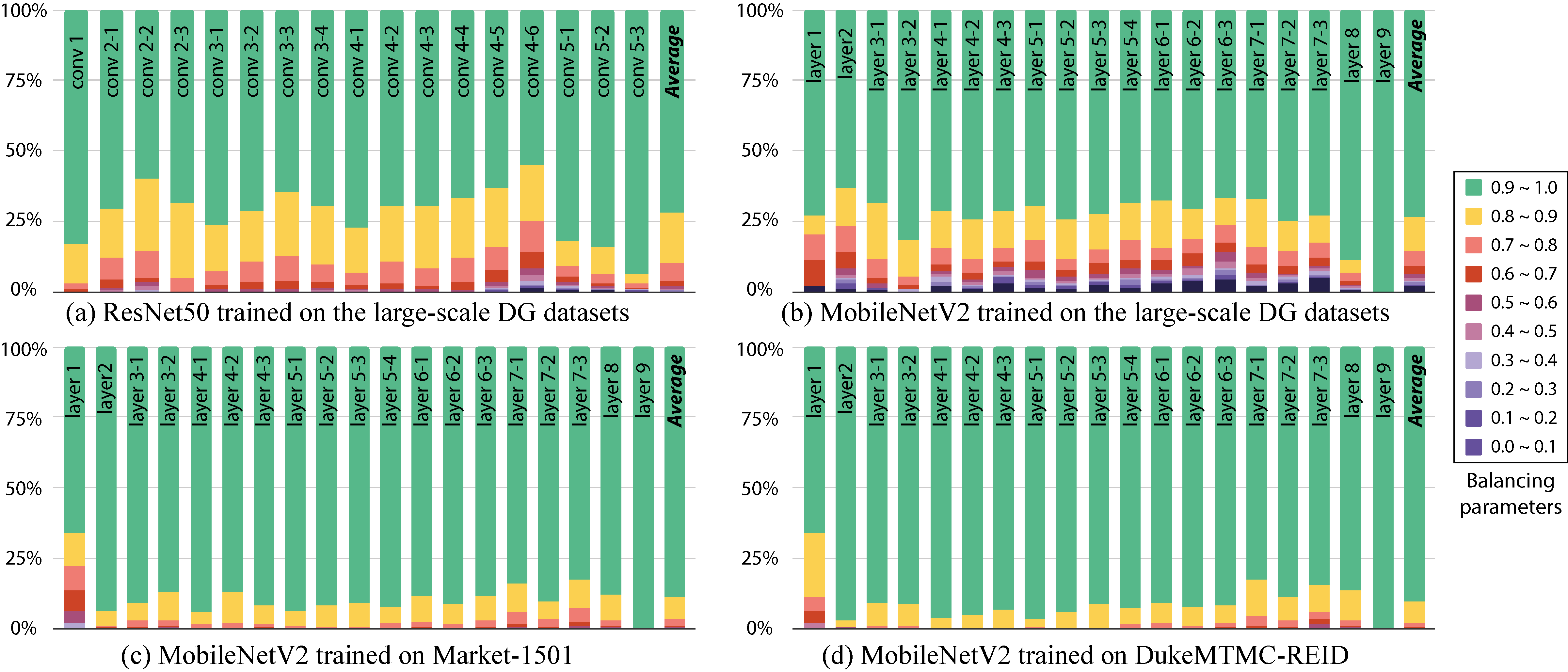}
\end{center}
\vspace{-0.20cm}
\caption{
Analysis on the final distribution ratios of balancing parameters according to the network structures (\ie MobileNetV2 and ResNet50) or training datasets (\ie Market-1501, DukeMTMC-ReID, and large-scale domain generalization datasets).
}
\vspace{-0.10cm}
\label{fig:bin_hist}
\end{figure*}

\vspace{3.0cm}

\section{More Analysis on Balancing Parameters}

\textbf{Different network structures and training datasets:} 
We compare the final distribution ratios of balancing parameters under different situations. Figure~\ref{fig:bin_hist} (a) and (b) show the results corresponding to network structures. While the existing BIN method~\cite{nam2018batch} normalizes instance-specific styles only in the shallow and deep layers (Fig. 4 in the manuscript), our MetaBIN method focuses on normalizing styles in the overall layers excluding the last layer, regardless of the network architecture. Figure~\ref{fig:bin_hist} (c) and (d) show the final distributions of balancing parameters trained on Market-1501~\cite{zheng2015scalable} and DukeMTMC-ReID~\cite{zheng2017unlabeled}, respectively. At this time, we consider camera domains within a single-source dataset as multiple-source domains. 
We observed that the balancing parameters are hardly biased towards IN. The reason is that the camera-domain discrepancy within the single-source dataset is relatively smaller than the domain discrepancy between multiple-source datasets in the large-scale DG benchmark.

\begin{figure}[t]
\begin{center}
\vspace{0.00cm}
\includegraphics[width=1.0\linewidth]{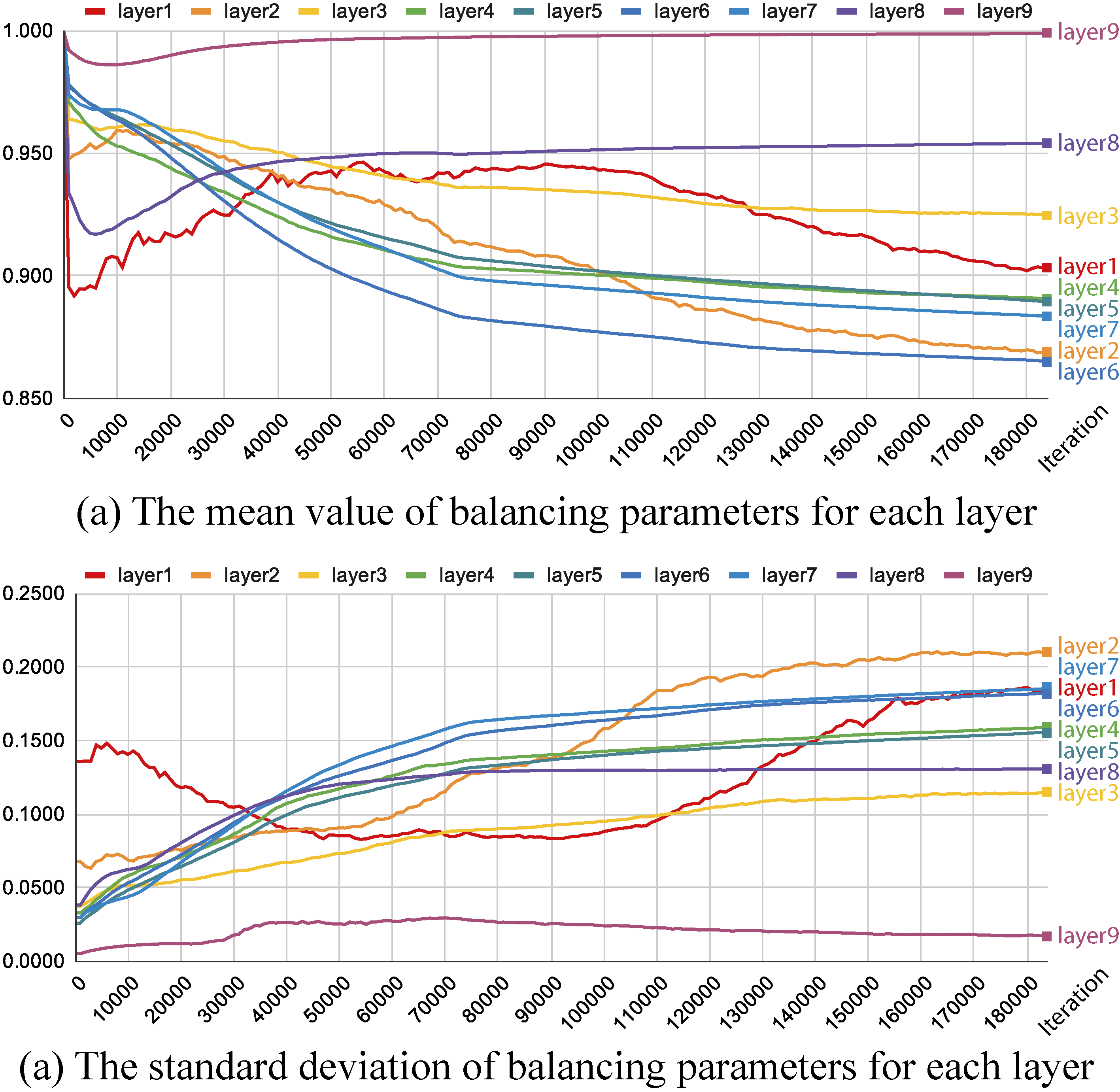}
\end{center}
\vspace{-0.10cm}
\caption{
Analysis on statistical characteristics of balancing parameters for each layer during the training process.
}
\vspace{-0.20cm}
\label{fig:bin_graph}
\end{figure}

\textbf{Parameter changes during training:} 
We analyze how the balancing parameters are updated during the training process, as shown in Fig. \ref{fig:bin_graph}. Interestingly, the mean value of balancing parameters in layers 1, 2, 8, and 9 dropped sharply at the beginning, but soon rebounded. It can be explained in line with the situation in Fig. \ref{fig:training_loss}. At the beginning of training, useless style information is removed from layers 1, 2, 8, and 9 to improve the discrimination ability rather than the generalization capability. After that, the balancing parameters are updated to overcome the unsuccessful generalization scenarios caused by the three types of meta train losses. Eventually, the model gradually becomes generalized through the training process.

\begin{table*}[tb]
	\caption{Performance (\%) and specification comparison under different network structures, where `Mem' is the training memory usage in our MetaBIN framework and `Time' is the inference time per image when the mini-batch size is 64.}
	\vspace{0.10cm}
	\centering
	\resizebox{\textwidth}{!}{
		\begin{tabular}{c||cc||cc||cc||cc||cc||c|c|c|c|c|c}
			 \Xhline{3\arrayrulewidth}
			 \multirow{3}{*}{Method} & \multicolumn{10}{c||}{Performance} & \multicolumn{6}{c}{Specification} \\ \cline{2-17} 
			  & \multicolumn{2}{c||}{Average}  & \multicolumn{2}{c||}{VIPeR} & \multicolumn{2}{c||}{PRID} & \multicolumn{2}{c||}{GRID} & \multicolumn{2}{c||}{i-LIDS} & \multirow{2}{*}{Dim} & Norm & Balancing & All & Mem & Time \\ \cline{2-11} 
			 & R-1 & \emph{m}AP & R-1 & \emph{m}AP & R-1 & \emph{m}AP & R-1 & \emph{m}AP & R-1 & \emph{m}AP && Layers & Params & Params & (MiB) & (ms) \\ \hline

MobileNetV2 (w1.0)      & 61.9 & 70.1 & 54.8 & 64.0 & 70.2 & 77.8 & 44.2 & 53.2 & 78.5 & 85.3 & 1,280 & 53  & 17,088 &  2.26M   & 5,907 & 1.34 \\ 
MobileNetV2 (w1.4)      & 64.7 & 72.3 & 56.9 & 66.0 & 72.5 & 79.8 & 49.7 & 58.1 & 79.7 & 85.5 & 1,792 & 53  & 23,822 &  4.34M   & 7,883 & 1.89 \\ \hline
ResNet18                & 59.5 & 67.5 & 54.2 & 63.0 & 65.0 & 74.2 & 42.1 & 49.8 & 76.7 & 83.0 & 512  & 20  & 4,800  &  11.19M  & 2,473 & 0.80 \\ 
ResNet34                & 62.8 & 71.0 & 57.1 & 66.0 & 73.0 & 79.5 & 43.7 & 54.1 & 77.3 & 84.4 & 512  & 36  & 8,512  &  21.30M  & 3,097 & 1.38 \\ 
ResNet50                & 66.0 & 73.6 & 59.9 & 68.6 & 74.2 & 81.0 & 48.4 & 57.9 & 81.3 & 87.0 & 2,048 & 53  & 26,560 &  23.56M  & 6,885 & 2.63 \\ 
ResNet101               & 68.1 & 75.9 & 61.5 & 70.2 & 77.1 & 83.3 & 52.7 & 62.8 & 81.2 & 87.2 & 2,048 & 104 & 52,672 &  42.61M  & 9,207 & 4.26 \\ 
ResNet152               & 68.3 & 75.8 & 62.4 & 70.7 & 74.1 & 81.9 & 53.3 & 61.9 & 83.5 & 88.8 & 2,048 & 155 & 75,712 &  58.30M  & 12,665& 5.92 \\ 

\Xhline{3\arrayrulewidth}
		\end{tabular}
	}
 \vspace{-0.00cm}
	\label{table:network}
\end{table*}

\section{Various architecture designs}

Domain generalizable person re-identification aims to learn a robust model for obtaining good performance on the unseen target domain without additional updates. This task is more useful for real-world applications since it does not require any target images to train a model. Therefore, we share experimental results on various network architectures for practical use. Especially, we cover the variants of MobileNetV2~\cite{sandler2018mobilenetv2} and ResNet~\cite{he2016deep}. Table \ref{table:network} shows the performance and specification corresponding to the different network structures. We employed a single NVIDIA Titan Xp GPU and set the input image size to $256 \times 128$. We also measured the maximum memory requirement for training the MetaBIN framework and calculated the inference time per image when a mini-batch was 64. We expect our MetaBIN method to be actively utilized in the real-world environment.

\end{document}